\documentclass[conference]{IEEEtran}
\IEEEoverridecommandlockouts
\usepackage{color, soul}
\usepackage{cite}
\usepackage{amsmath,amssymb,amsfonts}
\usepackage{algorithmic}
\usepackage{graphicx}
\usepackage{textcomp}
\usepackage{xcolor}

\def\BibTeX{{\rm B\kern-.05em{\sc i\kern-.025em b}\kern-.08em
    T\kern-.1667em\lower.7ex\hbox{E}\kern-.125emX}}
\begin{document}

\title{Herb–Drug Interactions: A Holistic Decision Support System in Healthcare \\
}

\author{\IEEEauthorblockN{Andreia Martins}
\IEEEauthorblockA{\textit{GECAD/ISEP} \\
\textit{Polytechnic of Porto}\\
Porto, Portugal \\
teles@isep.ipp.pt}
\and
\IEEEauthorblockN{Eva Maia}
\IEEEauthorblockA{\textit{GECAD/ISEP} \\
\textit{Polytechnic of Porto}\\
Porto, Portugal \\
egm@isep.ipp.pt}
\and
\IEEEauthorblockN{Isabel Praça}
\IEEEauthorblockA{\textit{GECAD/ISEP} \\
\textit{Polytechnic of Porto}\\
Porto, Portugal \\
icp@isep.ipp.pt}
}

\maketitle

\begin{abstract}
Complementary and alternative medicine are commonly used concomitantly with conventional medications leading to adverse drug reactions and even fatality in some cases. Furthermore, the vast possibility of herb-drug interactions prevents health professionals from remembering or manually searching them in a database. Decision support systems are a powerful tool that can be used to assist clinicians in making diagnostic and therapeutic decisions in patient care. Therefore, an original and hybrid decision support system was designed to identify herb-drug interactions, applying artificial intelligence techniques to identify new possible interactions. Different machine learning models will be used to strengthen the typical rules engine used in these cases. Thus, using the proposed system, the pharmacy community, people's  first  line  of  contact  within the Healthcare System, will be able to make better and more accurate therapeutic decisions and mitigate possible adverse events.

\end{abstract}


\section{Introduction}
Polypharmacy is an indisputable reality of the XXI century. In addition to the medications that doctors prescribe, patients can add a variety of over-the-counter herbs, supplements or even food which have a high potential to interact \cite{FUGHBERMAN2000134}. In addition, over the last few years, the usage of Complementary and Alternative Medicine (CAM), such as herbs and dietary supplements, has increased considerably~\cite{CHAVEZ20062146}. These products, unlike conventional drugs, embrace several bioactive entities that may hold therapeutic activity. CAM is very popular in several cultures and regions, and it is known by traditional or folk medicine, such as traditional Chinese, Tibetan, Japanese kampo, Indian ayurvedic and Yunani medicine~\cite{CHAVEZ20062146},~\cite{article1}.

Despite the benefits that CAM can bring at the therapeutic level, there are several reports in the literature of adverse events as a consequence of an Herb-Drug interaction (HDI) or a Supplement-Drug Interaction (SDI)~\cite{article1}. However, contrary to what happens with Drug-Drug Interactions (DDI) which is a challenge widely studied in clinical pharmacy, HDIs and SDIs are not a major concern in the pharmaceutical community~\cite{MALONE2004142, 10.1093/nar/gkab880}. Therefore, it becomes imperative to alert consumers, clinicians, pharmaceutical industries and health authorities, about the dangers of combining CAM with conventional drugs, as is already happening with drug combinations~\cite{FUGHBERMAN2000134, CHAVEZ20062146, article1}. 

\begin{figure*}[!]
\centering{\includegraphics[scale=.6]{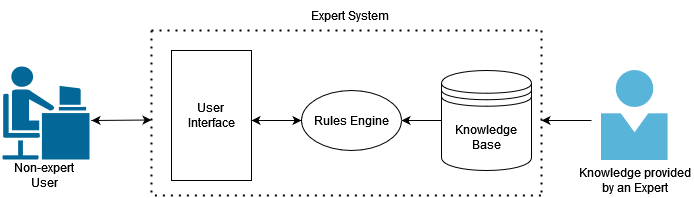}}
\caption{Basic structure of an expert system.}
\label{fig:expert}
\end{figure*} 

In medicine and healthcare, many computer systems have been developed to automatically assist clinicians and patients, such as medical experts systems~\cite{ZHOU202175}. An expert system (Fig.~\ref{fig:expert}), a sub-domain of Artificial Intelligence (AI), aggregates the knowledge provided by an expert into a knowledge base and encodes it as a set of \textit{if-then} rules, with an inference or rule engine, to emulate human thought processes so that the program operates at or near the level of human experts~\cite{article5}. The rule engine includes an explanation module that can show users how the system reached its conclusion. The user interface allows the user to get an answer to his question or problem in an intuitive and simplistic way. In particular, to deal with the mass of knowledge about DDIs, some works have applied this knowledge-based approach. For instance, an ancient study implemented an expert system to aid decision making in combination with drug therapy~\cite{PMID:3979039}, and another investigation resulted in a micro-computer based expert system on drug interactions~\cite{KINNEY1986462}. However, to the best of our knowledge, no other work has yet applied expert systems to the domain of HDIs~\cite{PMID:3979039,KINNEY1986462}.

Furthermore, recent studies have been applying other methods of AI to deal with DDIs, such as autoencoders and weighted Support Vector Machine (SVM)~\cite{article4}, Recurrent Neural Network (RNN)~\cite{10.1371/journal.pone.0190926}, among others. 
This range of investigations has the potential to improve the performance of an expert system. For example, some works suggest that, healthcare professionals have the ability to pay more attention to alerts that are most clinically significant if the volume of unnecessary DDI alerts is reduced~\cite{article3}. AI can help with this topic by selecting the ``right'' alerts to display and reducing the number of alerts healthcare workers are exposed to. Consequently, early investigations have been carried out to establish a list of high-priority DDIs for alerting purposes resorting to semi-supervised learning algorithms~\cite{article4}. This volume reduction would help to handle overwhelming amounts of data in the construction of the knowledge base of expert systems.

Based on previous studies and resources, ForPharmacy project~\cite{ForPharmacy} intends to research and develop telepharmacy solutions with particular attention to those in direct connection to pharmacovigilance and HDIs using pharmacies near patients to early detect and advise on distinct health-related risk factors. In particular, reliable tools that keep healthcare professionals up-to-date on potential HDIs are imperative. Therefore, in the context of the project, this article aims to present the conceptualization and contextualization of a Decision Support System (DSS) that will act as a critical tool to help local pharmacists to transform large amounts of clinical data into actionable knowledge to raise awareness about HDIs.

This work is organized into multiple sections that can be described as follows. Section II provides an overview on current state of the art of the HDIs domain and intelligent features that have emerged to address this problem. Section III describes the proposed system. Finally, Section IV provides a summary of the main conclusions of this work and appoints future research lines.

\section{Artificial Intelligence in CAM}
\label{sec:ai}

The worldwide popularity of herbal products have been incorporated into society healthcare supported by the perception that ``natural'' ensures safety~\cite{Zhang86},~\cite{Brantley301}. Furthermore, concomitant intake of herbal medicines and prescription drugs is a fairly common practice, particularly in patients with hypertension, diabetes, cancer, seizures, and depression~\cite{article7}. As a consequence, the risk of HDI is increasingly recognized as a public health problem~\cite{article7}. This problem can be accompanied by Adverse Drug Reactions (ADRs) that can lead to prolonged hospitalization and fatality in some cases~\cite{article7},~\cite{article6}. 

\begin{figure*}[!]
\centering{\includegraphics[scale=.55]{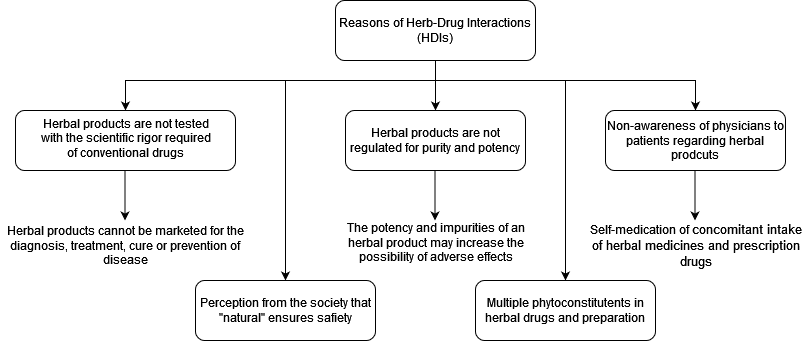}}
\caption{Main reasons of herb-drug interactions.}
\label{fig:intera}
\end{figure*}

\begin{figure*}[h!]
\centering{\includegraphics[scale=.55]{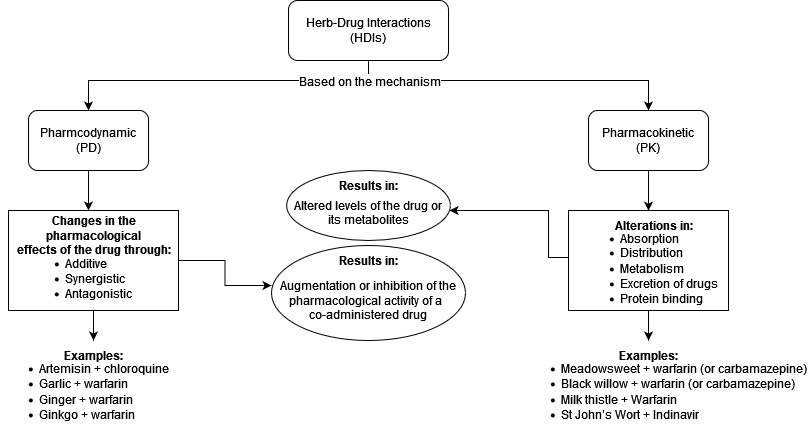}}
\caption{Main mechanisms of herb-drug interactions.}
\label{fig:pharma}
\end{figure*}

There are several reasons of herb-drug interactions. Fig.~\ref{fig:intera} tries to summarize the main reasons for these interactions~\cite{POLAKA2022,Oga2016}. Roughly speaking, HDIs can be characterized as Pharmacodynamic (PD) or Pharmacokinetic (PK), considering the mechanistic pathways through which the HDIs occur, resulting in null, beneficial or toxic responses.  PD interactions can occur when the constituents of herbal products have synergistic or antagonistic activity respecting to the conventional drug. PK interactions result from changes in the absorption, distribution, metabolism or elimination of the conventional drug by the herbal product or other dietary supplements. Drugs with a narrow therapeutic index are, especially, a major security concern relative to potential HDIs. Warfarin, the most commonly used anticoagulant, has a narrow therapeutic index and a lot of medicinal herbs and food interactions~\cite{article15}. In the investigation carried out by C. Awortwe \textit{et al.}~\cite{article7}, it was concluded that, in most cases, patients using warfarin and/or statins (atorvastatin, simva statin and rosuvastatin) for the treatment of cardiovascular complications described interactions after the combination with herbal products such as sage, flaxseed, SJW, cranberry, goji juice, green tea and chamomile. Potential interaction of warfarin and active constituents of herbal products led to ADRs including ecchymosis, epistaxis, haematuria, hemiplegia and elevated INR~\cite{article7}. However, despite the negative consequences that various HDIs can bring, others have a beneficial effect on therapy when properly prescribed to the patient~\cite{CHAVEZ20062146}. Fig.~\ref{fig:pharma} presents these main mechanisms of HDIs~\cite{Oga2016, herbs}.

In the late 1990s, after realizing the clinical importance of HDIs, the scientific community and companies began to develop databases of different HDIs using Information Technology (IT). For instance, UW Drug Interaction Database (DIDB)~\cite{uw} is a commercially available HDI Database founded by Dr. René Levy at the University of Washington in the late 1990s. Although it was created several years ago, it is updated daily and manually validated by experts. In june 2021, DIDB contained a total of 2,539 natural products (herbal medications and food products), with 15,864 drug interaction experiments/studies~\cite{Zhang86}. 

In the last decade, AI technologies, especially Natural Language Processing (NLP), have been applied in building HDI databases. This computational technique is used to analyze and represent naturally occurring texts with the goal of achieving human-like language processing for several applications~\cite{nlp}. Thus, NLP is particularly relevant on this topic as it contributes to understand and organize large amounts of biomedical text data~\cite{article8},~\cite{8210822}. To date, the most representative example of the application of AI for an HDI database is SUPP.AI~\cite{Zhang86}, developed by L. Wang \textit{et al.}, in 2019~\cite{wang-etal-2020-supp}. This database provides evidence of SDIs by automatically extracting supplement information and recognize such interaction from the scientific literature. The authors applied the RoBERTa language model~\cite{article10}, an iteration of BERT~\cite{https://doi.org/10.48550/arxiv.1810.04805}, using labeled data for DDI classification. The aforementioned model, on the SDI test set, reached 82\% precision, 58\% recall and 68\% F1-score. Unlike the DIDB, that requires a manual curation effort, SUPP.AI is updated with no manual validation once in several months~\cite{Zhang86}. 

Moreover, an intuitive function interface is equally important to make full advantage of the collected data. Without it the end user will not be able to use the information gathered. Thus, databases designed to support healthcare professionals usually have the concern of developing some key interfaces to support the user, namely keyword-based searches, providing alphabetical indexes to facilitate searches and advanced search options in order to simplify filtration of undesirable search results. For example, Natural Medicines Comprehensive Database (NMCD), a collection of databases, allows users to eliminate certain fields for a particular query~\cite{Zhang86}. 

Despite AI's ability to process a large amount of data in order to achieve greater coverage of HDI information, there are still a number of limitations in database development. Some of the limitations are related to improve the accuracy of methods to perform various NLP tasks as well as some databases have not been updated for a few years. Furthermore, due to the rule-based and probabilistic methods used in database development, developers of HDI databases should plan ahead to carry out extensive method testing before implementing them~\cite{Zhang86}.

Another problem is associated to the search for articles related to HDIs. Some studies suggested that searches conducted by physicians just achieve between 31-46\% of relevant articles~\cite{article12}. To overcome this disadvantage, K. Lin \textit{et al.}~\cite{article11} construct an automated HDI interaction PubMed-based article retrieval system. This system avoid completely the need for users to write a PubMed query, accepting simple medication and herb names as input and returning the articles that are relevant. Evaluation was based on a randomly selected set of herb-drug pairs from a previous review article~\cite{article13}, achieving a precision score of 93\% and a sensitivity score of 92\%.

In~\cite{inproceedings}, D. Trinh \textit{et al.}, proposed an approach for semantic relation clustering with the aim of extracting potential HDIs from the biomedical literature and, consequently, saving time in such investigations. The authors used a feature reduction method, Principal Component Analysis (PCA), to perform sparse feature reduction and applied K-means to cluster all the potential relations. The system reached 54.45\% precision, 75.71\% recall and 63\% F-score. 

Several developments in the HDI domain have occurred recently, however more advanced computational methods are needed to provide a better understanding of CAM to physicians and the public in general~\cite{WinNT}. In that regard, some methods based on AI that have been already proposed for DDIs are also promising to be applied to the context of HDIs. N. Lui \textit{et al.}, proposed a Machine Learning (ML) framework with the aim of extracting useful features from the Food and Drug Administration (FDA) adverse event reports and, afterwards, identify potential high-priority DDIs. The introduced approach combines Stacked Autoencoders (SAE) and Weighted Support Vector Machine (wSVM). The experimental results demonstrate the profitable performance of the proposed algorithm to predict high-priority DDI candidates for medication alerts as well as that features derived from adverse events contains effective information regarding severity levels of DDIs~\cite{article4}.

In~\cite{10.1371/journal.pone.0190926}, S. Lim \textit{et al.}, used a RNN to improve the performance of DDI extraction in the biomedical domain. The proposed model overcame the state-of-the-art model by 4.4\% and 2.8\% in the detection and classification tasks, respectively.

Expert systems have also been applied in DDI context, since it is crucial to deal with the large amount of knowledge about DDIs to the individual patient. In~\cite{article2}, A. Mahdi \textit{et al.}, introduced an expert system that draw conclusions from more complex interactions, using cat swarm algorithm~\cite{article14} to get accurate and fast results. This system can identify possible interactions between drugs, drugs and disease´s case and between drugs and food along with alternative drug suggestions that can be safer.
However, to the best of our knowledge, no other work has yet applied expert systems to the domain of HDIs~\cite{PMID:3979039},~\cite{KINNEY1986462}. There are several advantages of the rule-based reasoning approach such as natural expression~\cite{Durkin1994ExpertS}. Often humans tend to express their knowledge in \textit{if-then} rules, similar to the way knowledge is encoded into expert systems. Another advantage is related to the fact that rules are independent pieces of knowledge and consequently, can be easily reviewed and verified by experts. Furthermore, the rules are more transparent while compared with other forms of knowledge representation, for example, those employing neural networks~\cite{liebowitz2019handbook}. 

\section{ForPharmacy Approach}

\begin{figure*}[h!]
\centering{\includegraphics[scale=.6]{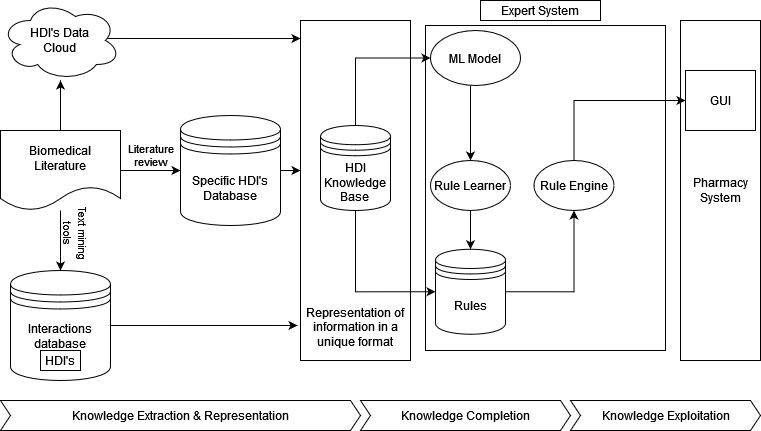}}
\caption{ForPharmacy approach for constructing a decision support system of herb-drug interactions.}
\label{fig}
\end{figure*} 

Most CAM products do not require a prescription and can be purchased at pharmacies or para-pharmacies. Therefore, pharmacists are the most suitable professionals to early detect possible HDIs, educate consumers about the use of CAM and, consequently, decrease the risk of HDIs, promoting the health of general population~\cite{Popattia2021ImprovingPP}. By this means, reliable tools that keep health professionals updated and help them obtain information quickly and usefully are required. The goal of ForPharmacy project is to expand access to pharmaceutical care by providing greater patient safety and counseling. Therefore, in the scope of the project, it is intended to develop an intelligent service to provide awareness about HDIs, particularly in relation to self-medication. 

To accomplish the aforementioned, Fig.~\ref{fig} shows a draft of the proposed system. Three different general phases can be identified as follows:

\begin{itemize}
    \item \textbf{Knowledge extraction and representation:} the information about HDIs is extracted from different sources, e.g., directly from biomedical literature or using HDIs databases. As already mentioned in Section II this can be a huge task taking into account that the information is not in a standard format. Although, AI techniques, namely NLP can be used to automate this step. 
    \item \textbf{Knowledge completion:} the extracted information needs must be standardized to be easily integrated in the expert system.
    \item \textbf{Knowledge exploitation:} the standardized information needs to be analysed and correlated to exploit and provide new knowledge to the pharmaceutical system.
\end{itemize}

These three phases are crucial to ensure the efficiency of the system. Each phase has different components and brings different challenges to the development of the intelligent service. In the following paragraphs, we briefly describe each phase. 

A methodology for extracting HDIs from textual data is strongly needed since the inability of non-experts to review the biomedical literature about potential HDIs is one of the causes that lead to the combination of herb-based products and prescription drugs~\cite{inproceedings}. Therefore, knowledge extraction and representation is the first phase and can be achieved through different approaches. An approach involves conducting a literature review on the subject of HDIs. Subsequently, an evidence-based approach is needed to evaluate the interactions. This evaluation must take into account several parameters, for a greater standardization of the data, such as reported by M. Chavez \textit{et al.}~\cite{CHAVEZ20062146}, and C. Awortwe \textit{et al.}~\cite{article7}. On the other hand, and considering that manually organizing information in a database is a costly and time consuming process, text mining tools, already applied for DDIs as mentioned in Section~\ref{sec:ai}, can be applied to automatically extract the information. Finally, the development of a data cloud is another approach to organize and integrate all the information from the existing HDI databases already mentioned in Section II. Any of these approaches is scalable and can be combined with other themes, such as DDIs and SDIs. 

All these sources should be consider when gathering data for the knowledge base. Moreover, it is essential to work the information provided by the different sources and represent it in a unique, standard format. In this way, it becomes possible to define standards to fill the information and, subsequently, serve as input into the expert system.  This is done in the second phase designated by knowledge completion. Note that, as we have multiple sources, it is also important to do a first analysis of the information to ensure that we do not duplicate information from different sources.

Then, all the knowledge gathered from different sources and represented in the standard format will be used by the expert system to exploit and provide new knowledge. This is the ``heart'' of ForPharmacy DSS system, and is developed in the final phase named by knowledge exploitation. The proposed expert system will work in a hybrid mode, not only the typical rule engine will be implemented, but it will be empowered by ML models. Thus, the acquired knowledge is encoded in a set of \textit{if-then} rules in order to maintain the explainability of HDIs. The rule engine runs an inference engine to process the rules in the system to identify possible interactions. 

To generate new rules from the knowledge base, different ML techniques will be implemented. Thus, the rule learner will use the best ML models to generate new rules and inserts them into the rule system. Several ML techniques will be studied to develop the rule learner, however one of the ML techniques that seems to be most suitable for this case is the Decision Tree, a supervised learning technique that is a very simple and widely used algorithm, since it offers a very good trade-off between performance and interpretability, favouring the latter~\cite{mlbook}. 

As already mentioned in Section~\ref{sec:ai}, it is extremely important to ensure that the knowledge extracted is presented to the health user in the simplest and most accessible way. Also, in more specific industries such as healthcare, maintaining consistency with the user interface of the already used systems is crucial to ensure that users do not have a learning curve to use the new system. ForPharmacy DSS system will have a GUI that will be integrated in the pharmaceutical proprietary platform used in most Portuguese pharmacies. The interface design will be in accordance with the system that pharmacists are familiar with and will allow them to select the drug and herb names and verify if there is any interaction. If a possible interaction is identified, the information provided includes explanations of what may happen when two specific herbs and drugs are used together and why this interaction may occur.

\section{Conclusion}
ForPharmacy project aims to develop technologies that enable pharmacies to offer a wider and more reliable range of healthcare services. In particular, to the best of our knowledge pharmacies do not own multidisciplinary tools to alert the pharmacist about the potential risk to customers related to HDIs. In the scope of ForPharmacy project we aim to demonstrate for the first time, a DSS for HDIs which uses AI techniques to identify new interactions.
It will work in an innovative hybrid mode: not only the typical rule engine will be implemented, but it will be empowered by ML models. Therefore, this system, which is currently under development, will facilitate the daily life of pharmacists, helping them in preventing possible HDIs caused by self-medication. 
Furthermore, it should be noted that this new system is fully scalable to other subjects, such as DDIs and SDIs, which can greatly improve pharmacy systems. 

\section*{Acknowledgment}
The present work was done and funded under project ForPharmacy (P2020-COMPETE-FEDER nr 070053). This work has also received funding from project UIDB/00760/2020.

\bibliographystyle{unsrt}
\bibliography{bibliography}

\begin{thebibliography}{10}

\bibitem{FUGHBERMAN2000134}
Adriane Fugh-Berman.
\newblock Herb-drug interactions.
\newblock {\em The Lancet}, 355(9198):134--138, 2000.

\bibitem{CHAVEZ20062146}
Mary~L. Chavez, Melanie~A. Jordan, and Pedro~I. Chavez.
\newblock Evidence-based drug–herbal interactions.
\newblock {\em Life Sciences}, 78(18):2146--2157, 2006.
\newblock NATURECEUTICALS (NATURAL PRODUCTS), NUTRACEUTICALS, HERBAL
  BOTANICALS, AND PSYCHOACTIVES: DRUG DISCOVERY AND DRUG-DRUG INTERACTIONS.

\bibitem{article1}
Khalid Parvez and Vikas Rishi.
\newblock Herb-drug interactions and hepatotoxicity.
\newblock {\em Current Drug Metabolism}, 20, 03 2019.

\bibitem{MALONE2004142}
Daniel~C. Malone, Edward~P. Armstrong, Jacob Abarca, Amy~J. Grizzle, Philip~D.
  Hansten, Robin~C. {Van Bergen}, Babette~S. Duncan-Edgar, Steven~L. Solomon,
  and Richard~B. Lipton.
\newblock Identification of serious drug–drug interactions: Results of the
  partnership to prevent drug–drug interactions.
\newblock {\em Journal of the American Pharmacists Association},
  44(2):142--151, 2004.

\bibitem{10.1093/nar/gkab880}
Guoli Xiong, Zhijiang Yang, Jiacai Yi, Ningning Wang, Lei Wang, Huimin Zhu,
  Chengkun Wu, Aiping Lu, Xiang Chen, Shao Liu, Tingjun Hou, and Dongsheng Cao.
\newblock {DDInter: an online drug–drug interaction database towards
  improving clinical decision-making and patient safety}.
\newblock {\em Nucleic Acids Research}, 50(D1):D1200--D1207, 10 2021.

\bibitem{ZHOU202175}
Li~Zhou and Margarita Sordo.
\newblock Chapter 5 - expert systems in medicine.
\newblock pages 75--100, 2021.

\bibitem{article5}
Richard~O. Duda and Edward~H. Shortliffe.
\newblock Expert systems research.
\newblock {\em Science}, 220(4594):261--268, 1983.

\bibitem{PMID:3979039}
J~Roach, S~Lee, J~Wilcke, and M~Ehrich.
\newblock An expert system for information on pharmacology and drug
  interactions.
\newblock {\em Computers in biology and medicine}, 15(1):11—23, 1985.

\bibitem{KINNEY1986462}
Evlin~L. Kinney.
\newblock Expert system detection of drug interactions: Results in consecutive
  inpatients.
\newblock {\em Computers and Biomedical Research}, 19(5):462--467, 1986.

\bibitem{article4}
Ning Liu, Cheng-Bang Chen, and Soundar Kumara.
\newblock Semi-supervised learning algorithm for identifying high-priority
  drug-drug interactions through adverse event reports.
\newblock {\em IEEE Journal of Biomedical and Health Informatics}, PP:1--1, 08
  2019.

\bibitem{10.1371/journal.pone.0190926}
Sangrak Lim, Kyubum Lee, and Jaewoo Kang.
\newblock Drug drug interaction extraction from the literature using a
  recursive neural network.
\newblock {\em PLOS ONE}, 13(1):1--17, 01 2018.

\bibitem{article3}
Shobha Phansalkar, Heleen Sijs, Alisha Tucker, Amrita Desai, Douglas Bell,
  Jonathan Teich, Blackford Middleton, and David Bates.
\newblock Drug-drug interactions that should be non-interruptive in order to
  reduce alert fatigue in electronic health records.
\newblock {\em Journal of the American Medical Informatics Association :
  JAMIA}, 20, 09 2012.

\bibitem{ForPharmacy}
Forpharmacy.
\newblock https://inovglintt.com/projetos/forpharmacy/.
\newblock Accessed: 2022-07-26.

\bibitem{Zhang86}
Yufeng Zhang, Chung Man~Ip, Yuen~Sze Lai, and Zhong Zuo.
\newblock Overview of current herb{\textendash}drug interaction databases.
\newblock {\em Drug Metabolism and Disposition}, 50(1):86--94, 2022.

\bibitem{Brantley301}
Scott~J. Brantley, Aneesh~A. Argikar, Yvonne~S. Lin, Swati Nagar, and Mary~F.
  Paine.
\newblock Herb{\textendash}drug interactions: Challenges and opportunities for
  improved predictions.
\newblock {\em Drug Metabolism and Disposition}, 42(3):301--317, 2014.

\bibitem{article7}
Charles Awortwe, Memela Makiwane, Helmuth Reuter, Christo Muller, Johan Louw,
  and Bernd Rosenkranz.
\newblock Critical evaluation of causality assessment of herb‐drug
  interactions in patients.
\newblock {\em British Journal of Clinical Pharmacology}, 84, 01 2018.

\bibitem{article6}
Martins Ekor.
\newblock The growing use of herbal medicines: Issues relating to adverse
  reactions and challenges in monitoring safety.
\newblock {\em Frontiers in pharmacology}, 4:177, 01 2014.

\bibitem{POLAKA2022}
Suryanarayana Polaka, Sayali Chaudhari, Muktika Tekade, Mukesh~Chandra Sharma,
  Neelesh Malviya, Sapna Malviya, and Rakesh~Kumar Tekade.
\newblock Chapter 13 - clinical importance of herb–drug interaction.
\newblock In Rakesh~Kumar Tekade, editor, {\em Pharmacokinetics and
  Toxicokinetic Considerations}, volume~2 of {\em Advances in Pharmaceutical
  Product Development and Research}, pages 323--356. Academic Press, 2022.

\bibitem{Oga2016}
Enoche~F. Oga, Shuichi Sekine, Yoshihisa Shitara, and Toshiharu Horie.
\newblock Pharmacokinetic herb-drug interactions: Insight into mechanisms and
  consequences.
\newblock {\em European Journal of Drug Metabolism and Pharmacokinetics},
  41(2):93--108, Apr 2016.

\bibitem{article15}
Natasa Milic, Nataša Milošević, Svetlana Golocorbin-Kon, Bozic Teodora,
  Ludovico Abenavoli, and Francesca Borrelli.
\newblock Warfarin interactions with medicinal herbs.
\newblock {\em Natural product communications}, 9:1211--1216, 09 2014.

\bibitem{herbs}
Prasad~KVSRG Mamindla~S and Koganti B.
\newblock Herb-drug interactions: An overview of mechanisms and clinical
  aspects.
\newblock {\em International Journal of Pharmaceutical Sciences and Research},
  09 2016.

\bibitem{uw}
Drug interaction solutions.
\newblock https://www.druginteractionsolutions.org/, Jun 2022.

\bibitem{nlp}
S.~Meera and S.~Geerthik.
\newblock {\em Natural Language Processing}, chapter~10, pages 139--153.
\newblock John Wiley \& Sons, Ltd, 2022.

\bibitem{article8}
Raul Rodriguez-Esteban.
\newblock Biomedical text mining and its applications.
\newblock {\em PLoS computational biology}, 5:e1000597, 12 2009.

\bibitem{8210822}
Dan Xie, Wei Pei, Weiwei Zhu, and Xiaodong Li.
\newblock Traditional chinese medicine prescription mining based on abstract
  text.
\newblock In {\em 2017 IEEE 19th International Conference on e-Health
  Networking, Applications and Services (Healthcom)}, pages 1--5, 2017.

\bibitem{wang-etal-2020-supp}
Lucy Wang, Oyvind Tafjord, Arman Cohan, Sarthak Jain, Sam Skjonsberg, Carissa
  Schoenick, Nick Botner, and Waleed Ammar.
\newblock {SUPP}.{AI}: finding evidence for supplement-drug interactions.
\newblock pages 362--371, July 2020.

\bibitem{article10}
Yinhan Liu, Myle Ott, Naman Goyal, Jingfei Du, Mandar Joshi, Danqi Chen, Omer
  Levy, Mike Lewis, Luke Zettlemoyer, and Veselin Stoyanov.
\newblock Roberta: A robustly optimized bert pretraining approach.
\newblock 2019.

\bibitem{https://doi.org/10.48550/arxiv.1810.04805}
Jacob Devlin, Ming-Wei Chang, Kenton Lee, and Kristina Toutanova.
\newblock Bert: Pre-training of deep bidirectional transformers for language
  understanding.
\newblock 2018.

\bibitem{article12}
Salimah Shariff, Jessica Sontrop, and bhaynes@mcmaster.ca Haynes.
\newblock Impact of pubmed search filters on the retrieval of evidence by
  physicians (vol 184, pg e184, 2012).
\newblock {\em Canadian Medical Association Journal}, 184:1394--1394, 09 2012.

\bibitem{article11}
Kuo Lin, Carol Friedman, and Joseph Finkelstein.
\newblock An automated system for retrieving herb-drug interaction related
  articles from medline.
\newblock {\em AMIA Joint Summits on Translational Science proceedings. AMIA
  Summit on Translational Science}, 2016:140--9, 07 2016.

\bibitem{article13}
Angelo Izzo and Edzard Ernst.
\newblock Interactions between herbal medicines and prescribed drugs an updated
  systematic review.
\newblock {\em Drugs}, 69:1777--98, 02 2009.

\bibitem{inproceedings}
Khang Trinh, Duy Pham, and Ly~Le.
\newblock Semantic relation extraction for herb-drug interactions from the
  biomedical literature using an unsupervised learning approach.
\newblock pages 334--337, 10 2018.

\bibitem{WinNT}
Knowledge graph completion using artificial neural networks for herb-drug
  interaction discovery.
\newblock https://cordis.europa.eu/project/id/800578.
\newblock Accessed: 2022-06-10.

\bibitem{article2}
Amir Yasseen.
\newblock Design and implementation of rule-base expert system in drug
  interactions and support medical decision using cat swarm optimization
  algorithm.
\newblock 10:237--243, 01 2018.

\bibitem{article14}
Shu-Chuan Chu, Pei-Wei Tsai, and Jeng-Shyang Pan.
\newblock Cat swarm optimization.
\newblock pages 854--858, 01 2006.

\bibitem{Durkin1994ExpertS}
John Durkin.
\newblock Expert systems - design and development.
\newblock 1994.

\bibitem{liebowitz2019handbook}
Jay Liebowitz.
\newblock {\em The handbook of applied expert systems}.
\newblock cRc Press, 2019.

\bibitem{Popattia2021ImprovingPP}
Amber~Salman Popattia, Laetitia Hattingh, and Adam~La Caze.
\newblock Improving pharmacy practice in relation to complementary medicines: a
  qualitative study evaluating the acceptability and feasibility of a new
  ethical framework in australia.
\newblock {\em BMC Medical Ethics}, 22, 2021.

\bibitem{mlbook}
Trevor Hastie, Robert Tibshirani, and Jerome Friedman.
\newblock {\em The elements of statistical learning: data mining, inference and
  prediction}.
\newblock Springer, 2 edition, 2009.

\end{thebibliography}

\end{document}